# Image resolution enhancement for advanced semiconductor nodes

L. Rencker[1], O. Tajalizadehkhoob[1], K. Elsayed[1], A. Khachaturiants[1], H. Pahlavani[1], Y. Guo[1], J. van de Laar[1], E. Simons[1], N. Saikumar[1], H. Sadeghian[1]
[1]Nearfield Instruments, B.V. (The Netherlands)

**ABSTRACT**

Advanced semiconductor nodes are pushing the limits of feature sizes and require metrology with sub-nm resolution without compromising on the throughput as needed for in-line process control. Recently, high-throughput scanning probe microscopy (SPM) based metrology and inspection tools capable of meeting these needs have been introduced to the market and qualified for use in HVM. While innovative measurement methods and tool architecture have allowed for a leap of improvement in throughput, the next step in further reducing imaging time can be obtained through the application of machine learning for enhancing the resolution of measured images for extraction of relevant parameters. In this work, we provide the general framework under which a neural network-based resolution enhancer is designed and used for SPM images. We showcase the effectiveness of this framework using measurements performed on Line/Space structures with a pitch of 200 nm. For the reusability of a pre-developed pre-trained model, we additionally leverage transfer learning and show that a new model for slightly differing structures can be re-trained and calibrated with a smaller data set of measurements performed on Line/Space structures with a pitch of 100 nm.

**Keywords:** Scanning Probe Microscopy, Image processing, Machine Learning, Resolution Enhancement

## INTRODUCTION

The next generations of semiconductor nodes are being defined in terms of Angstroms instead of nm pointing towards the aggressive push by the industry to improve the feature resolutions printed on wafers, especially with the introduction of high-NA EUV lithography systems into the market. The unfortunate consequence of these systems is the linked increase in stochastic effects which could potentially result in a loss of yield. Process control through metrology with sub-nm resolution becomes critical for these nodes where full 3D profile capture of the developed and etched structures and subsequent extraction of the relevant parameters is necessary. High-throughput scanning probe microscopy-based metrology and inspection tools like QUADRA recently introduced into the market enable this through the introduction of innovative measurement modes allowing for profile measurements of complex 3D structures and combining this with clever tool architectures leading to short imaging times.

Although the high-speed tools like QUADRA can now achieve up-to 1 MPx/s for surface defect inspection scenario and rival other tools in terms of speed for metrology, there is a constant push to further increase throughput especially when the increase in stochastic defects with the next-generation nodes is considered whereby small sampling across the wafer for process control might be insufficient. As we move towards an era of massive metrology where measurements at a large number of locations across the wafer is necessary, all avenues for further reducing imaging time need to be explored. With SPM-based measurements, increasing throughput can come at the cost of losing image resolution which directly impacts the accuracy and precision of the parameters extracted from the image.

With the recent ubiquitous application of ML/AI methods in fabs, neural network (NN) based methods can be used to train models quickly and enhance the resolution of the images whereby scanning speed can be further increased resulting in a reduction of the true pixels in the measured image, which is then compensated through ML-based resolution enhancer for parameter extraction. The concept of enhancing the resolution of SPM images has been noted in other instances [1], and notably in [2] where an ML model was trained for images obtained on DNA samples wherein extraction of geometric and mechanical properties was of prime interest. In this work, we provide the general framework through which this method can be applied for SPM images obtained from measurements on semiconductor structures, where Line/Space structures are chosen to showcase the potential improvements possible with the framework.

In addition, we extend this framework to enable the utilization of transfer learning whereby retraining the network for similar structures of differing dimensions would be possible with minimal training data without compromising on the accuracy or precision of the extracted parameters.

## SPM MEASUREMENTS OF SEMICONDUCTOR STRUCTURES

For the purpose of showcasing the outcome of the NN-based framework, we identified Line/Space structures of different dimensions to be measured at different speeds. For the first part of the study focusing on enhancing the image resolution and improving the accuracy of the extracted parameters from images captured, we measured Line/Space structures with a pitch of 100 nm and 50% duty-cycle. The most relevant parameters of interest (POI) for this structure were identified to be line height, line width, line edge roughness (LER) and line width roughness (LWR). The scanning speed during measurements was first optimized till the accuracy and precision of extracted parameters over 5 static repeat measurements were within the acceptable window of uncertainty. This scanning speed is referred to as the optimized nominal scanning speed. Following this, additional measurements were performed at the same location by increasing the scanning speed to 5 and 7 times the nominal speed and capturing 5 static repeats for both cases. Example images captured at the nominal speed and 5 times the nominal speed are shown in Figure 1, where a clear difference in the captured cross-section profile is visible.

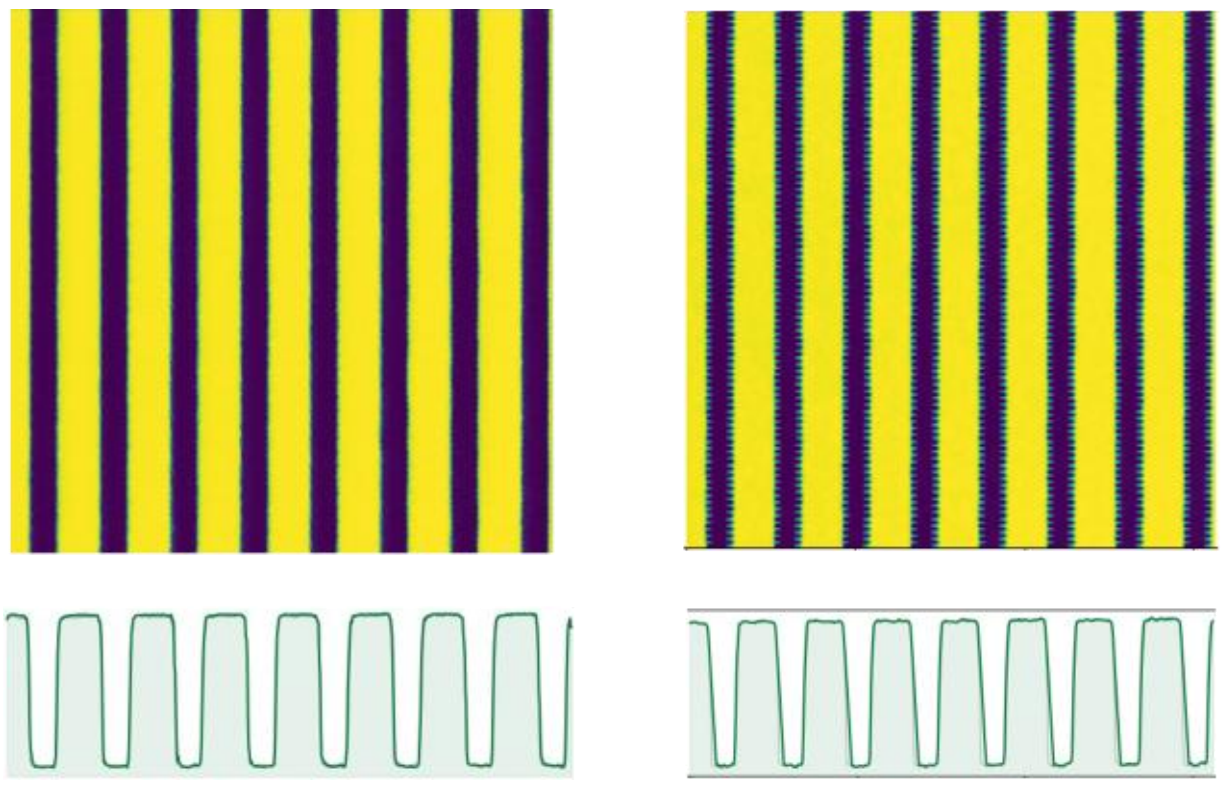

**Figure 1** Left: Optimized nominal speed high-resolution SPM image. Right: Low-resolution SPM image captured at 5x the nominal speed. Cross-section shows the change in profile measured which can be easily linked to the inaccuracy of extracted parameters.

All POIs noted above are extracted for all the measured images and assessed to understand the impact of increasing the scanning speed. The definition of the POIs is represented pictorially in Figure 2. The graphs from the extracted POIs are shown in Figures 3 and 4 respectively for POIs in the z-direction and lateral directions. From these extracted parameters, it is clear that for SPM measurement images, as long as at least a single pixel from the bottom of each Line/Space structure is part of the measured cross-section, a parameter like line height extracted in z-direction is unaffected. However, due to the resolution loss with increasing scanning speed, the lateral parameters extracted suffer heavily both in terms of accuracy and precision.

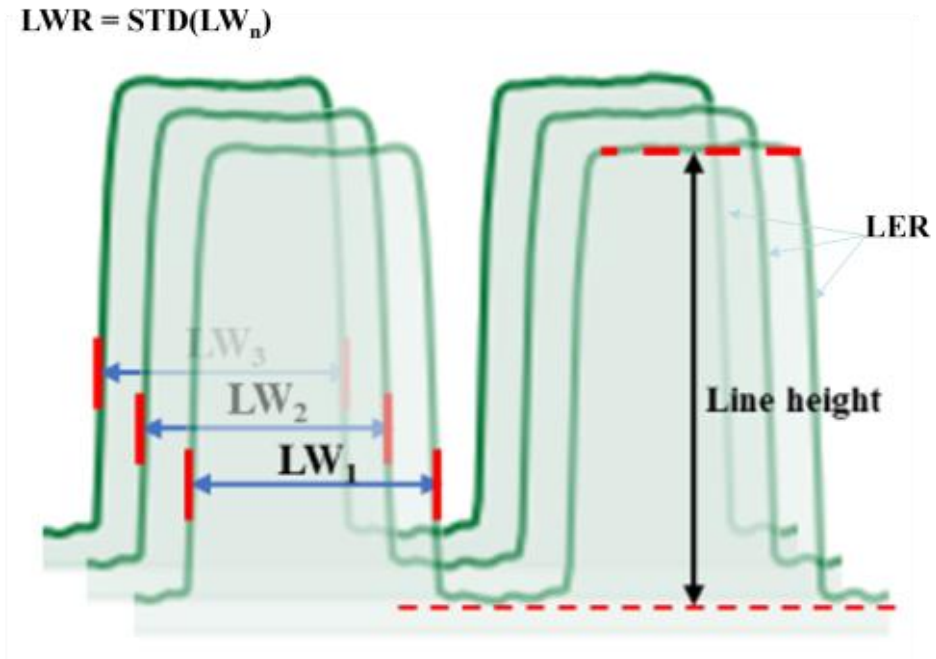

**Figure 2** Definition of extracted parameters of interest (POI)

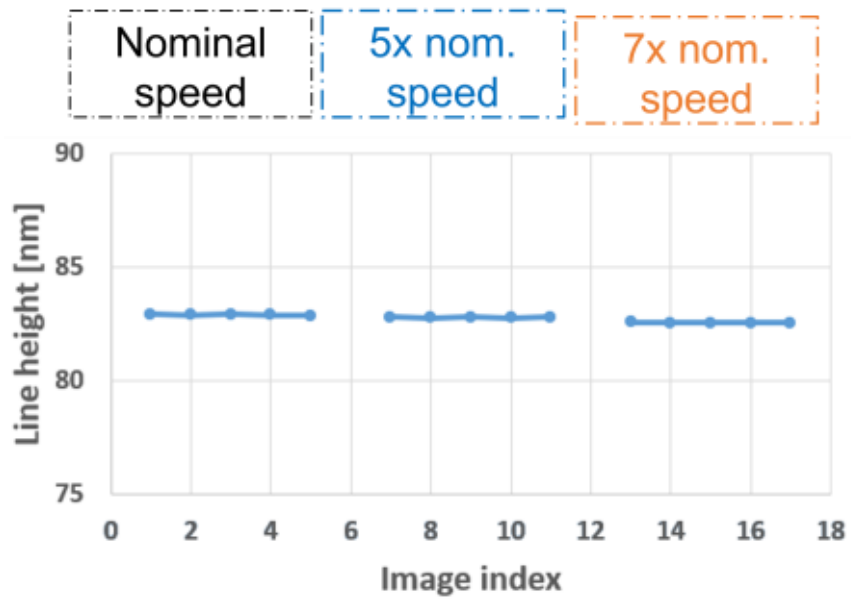

**Figure 3** Line height extracted at different speeds shows no change due to the increase of scanning speed and loss of pixel resolution

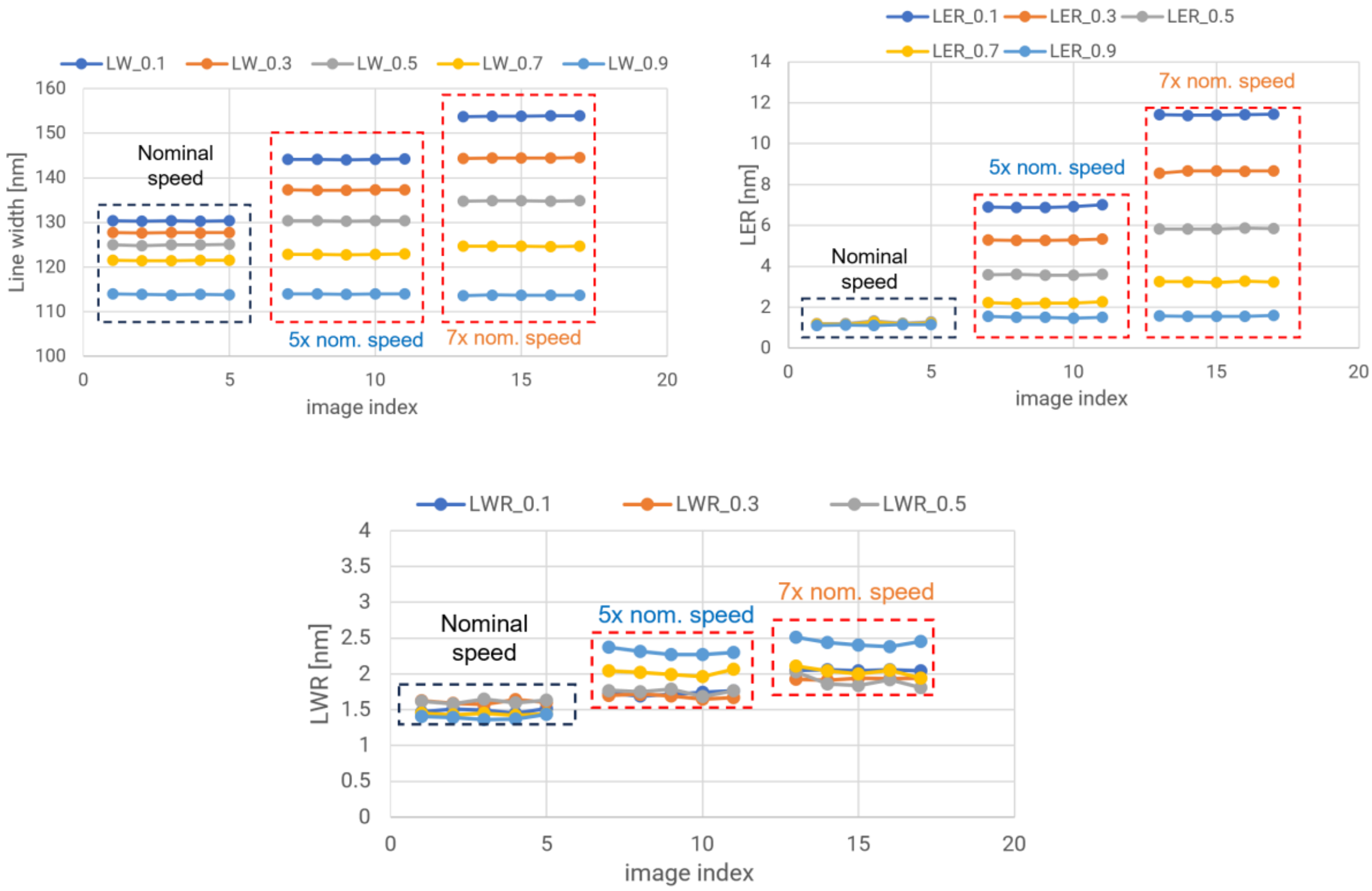

**Figure 4** Graphs showing the accuracy and precision of 3 different extracted POIs from the measured images, ie, Line width, Line edge roughness (LER) and Line width roughness (LWR). The subscript after the POI indicates the percentile height from the top of the line structure where these parameters are extracted from the full cross-section. The large inaccuracy caused by the increased scanning speed is clearly visible apart from the loss of precision which is also seen.

## ML FRAMEWORK FOR RESOLUTION ENHANCEMENT

In this work, we develop the ML framework using the Res-UNet which was introduced for image resolution enhancement, not aimed at SPM images [3] and has since then been extended to other applications [4]. This Res-UNet architecture utilizes deep learning like Convolutional Neural Networks (CNN) to learn the mapping from low towards high-resolution images and is widely used now in the fields of computer vision, medical imaging and also for sharpening satellite images. This architecture shown in Figure 5 is essentially a modification of the well-known U-Net architecture and introduces residual connections leveraging on the strengths of both for the image resolution enhancement task. The residual connections incorporated into the framework at multiple levels allow for the capture of both low and high-level features within the image, enabling the reconstruction of high-resolution details with great accuracy. It essentially utilizes the encoder-decoder structure for hierarchical feature extraction and propagation facilitating accurate reconstruction while at the same time preserving spatial information. In this work, as shown in Figure 5, we facilitate small filter sizes (3 x 3) and (2 x 2) to capture the necessary intricate details within the captured SPM images. The parameters from the model training are noted in Table 1.

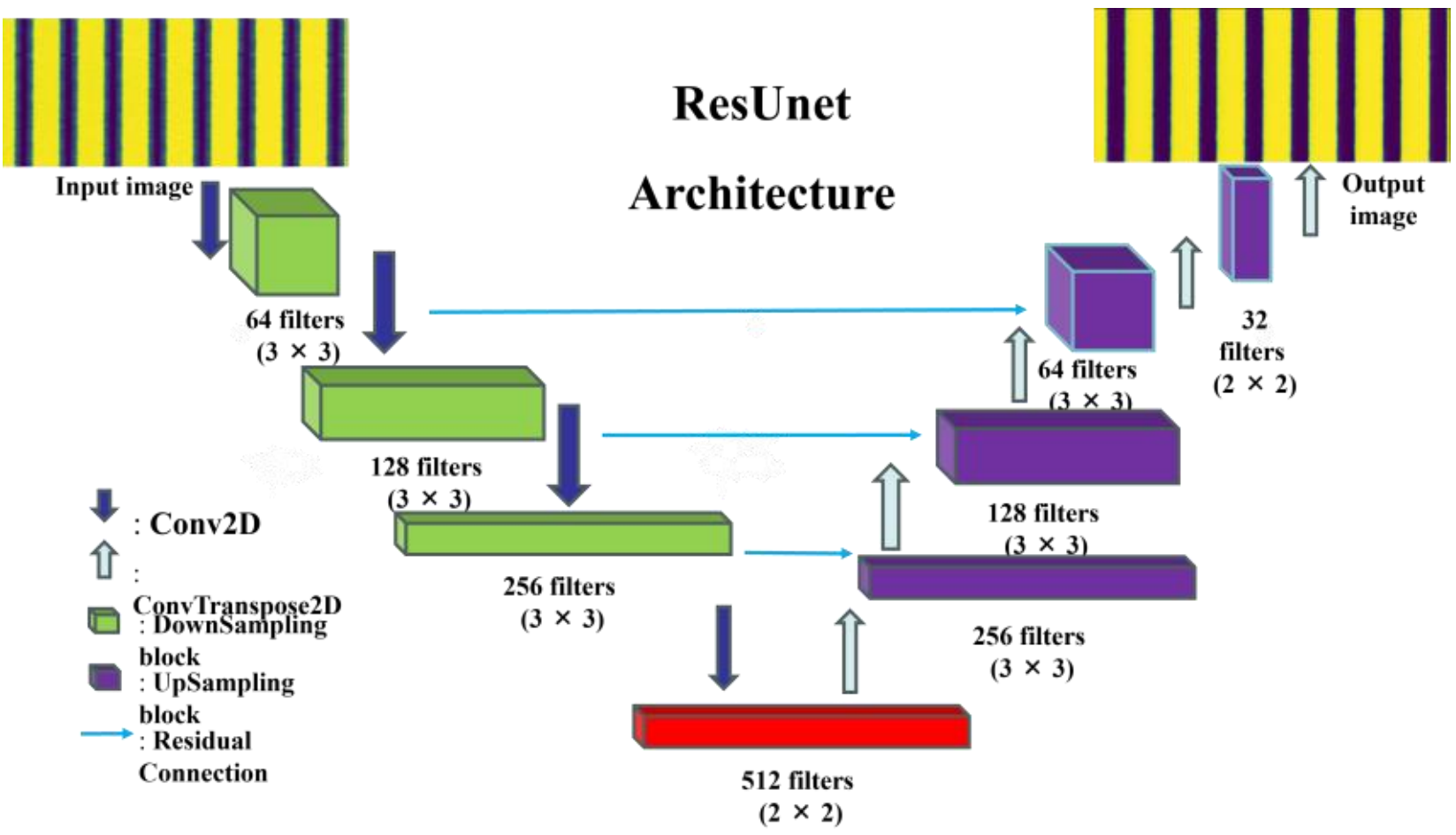

**Figure 5** Res-UNET architecture used for image resolution enhancement

| Parameter | Value |
|---|---|
| Input size | Any dimensions divisible by 16 |
| Learning rate | 1e-3 |
| Training loss | Mean Square error (MSE) |
| Output size | Same as input |
| Evaluation metric | MSE, PSNR |
| Activation Function (Output) | Sigmoid |
| Optimizer | Adam |
| Epochs | 250 |

**Table 1** Res-UNet based model training parameters

## RESULTS AND TRANSFER LEARNING

The results of the implementation of the trained ML model for resolution enhancement are shown pictorially in Figure 6, where from a look at the 2D image and the cross-section, the recovery of lost pixels is evident. More importantly, from the perspective of process control in fabs, the extracted parameters from the resolution-enhanced images is compared to the values obtained from the images measured at nominal speeds and the raw images obtained at higher than optimized nominal speeds in Figure 7 for all the lateral parameters. This clearly indicates that this approach for further enhancing the throughput of SPM-based metrology tools is successful.

A critical part necessary for the implementation of such a model in the fab is the ability to learn faster for differing dimensions found across different semiconductor layers being fabricated on the floor. One of the biggest issues with ML approaches is the necessity of large data sets and if a new ML model needs to be trained with a large number of images captured in each case, this creates an unnecessary hurdle. Here, we showcase the use of transfer learning where the same ML model which has been trained for the Line/Space structure of pitch 200 nm is retrained for structures of pitch 100 nm with a lower learning rate of 1e-4 to avoid overfitting and with a significantly reduced number of epochs of 50 for faster convergence and utilization in the lab. This strategy also allows for reducing the needed data set size by a factor of 4. The measurements on the 100 nm pitch structures are performed in a similar fashion with parameters extracted from the raw as well as resolution-enhanced images output by the ML model. These results are showcased in Figure 8 showing the effective utilization of transfer learning for this application.

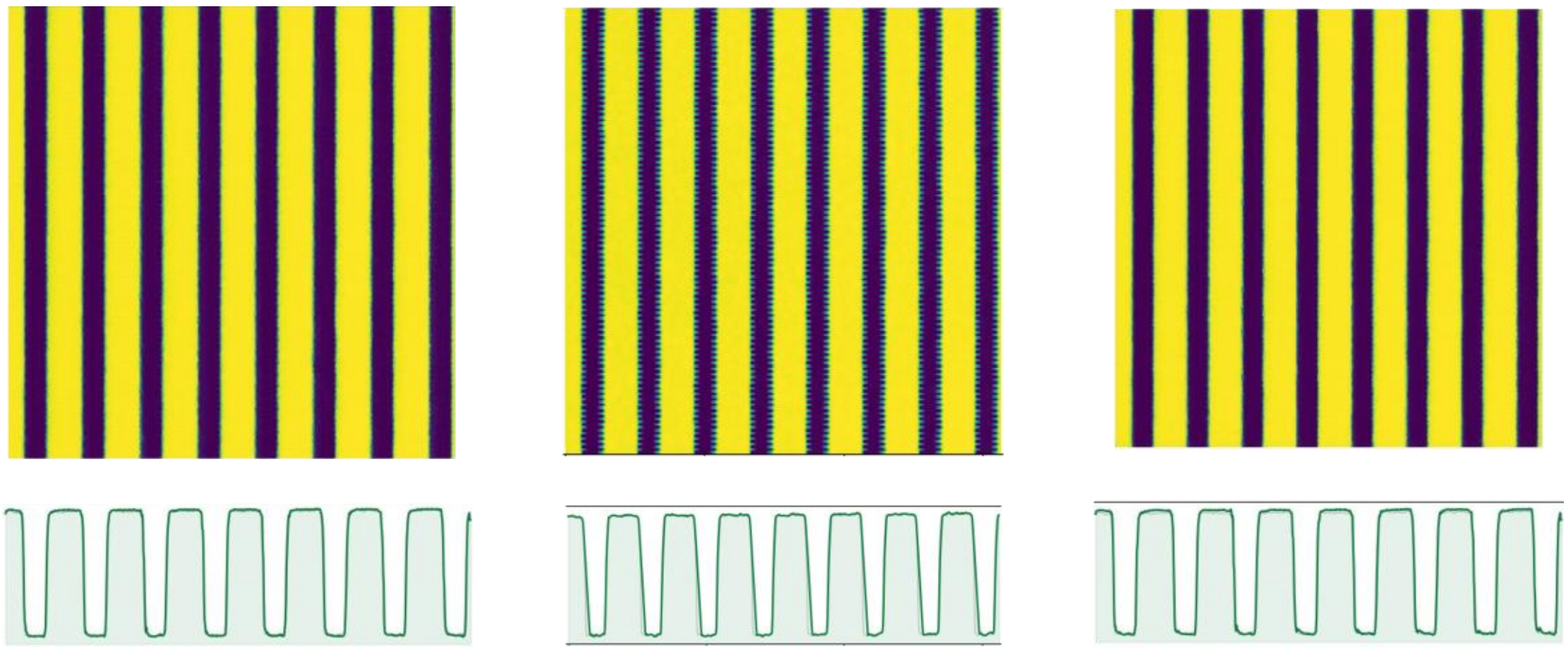

**Figure 6** SPM images with cross-sections shown. Left and middle image are repeats of the images shown in Figure 1. The middle and right images are the input and output images of the ML model.

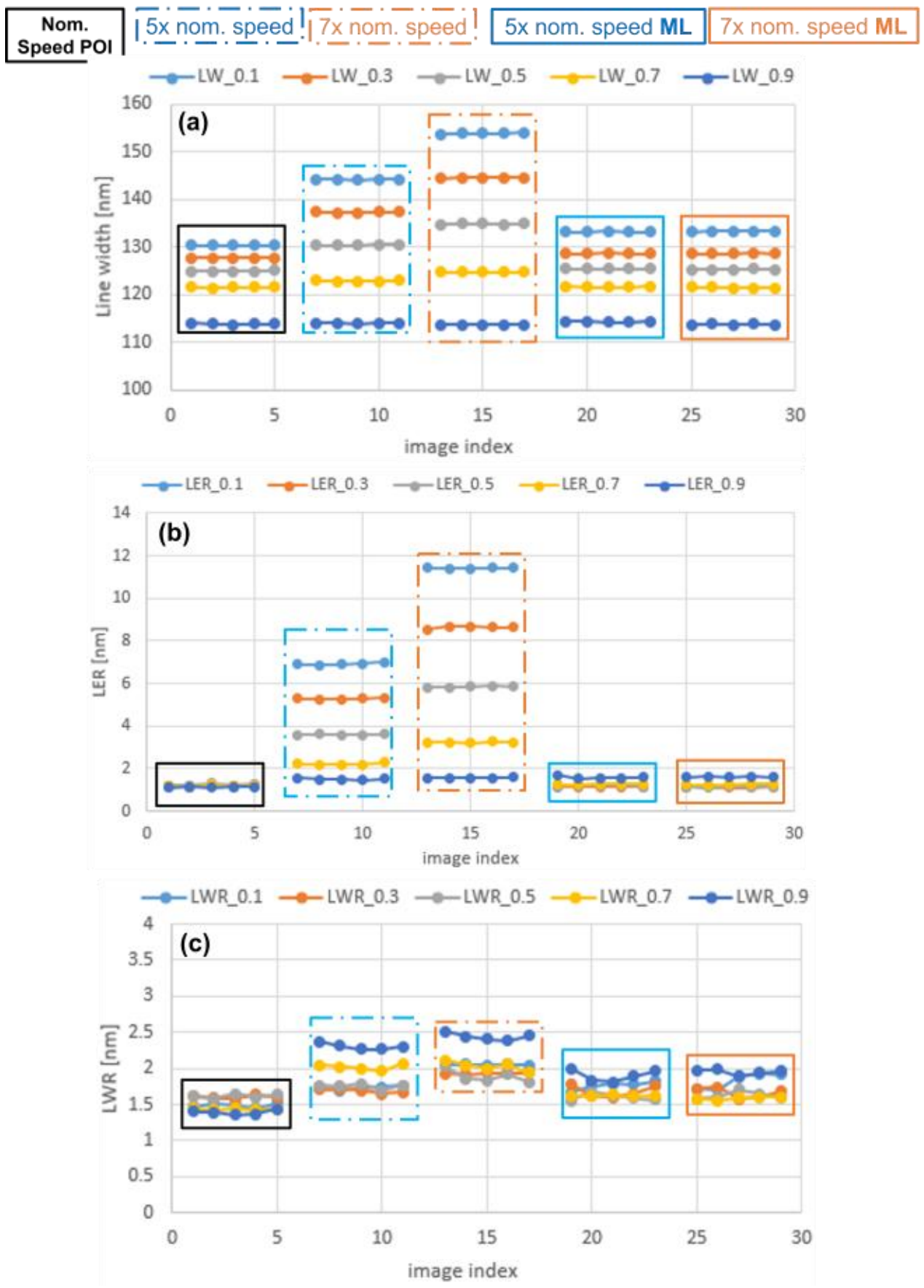

**Figure 7** Graphs showing the accuracy and precision of the same 3 extracted lateral POIs as in Figure 4 from the measured images including those extracted from the images where the trained Res-UNet ML model has been used to produce resolution-enhanced images

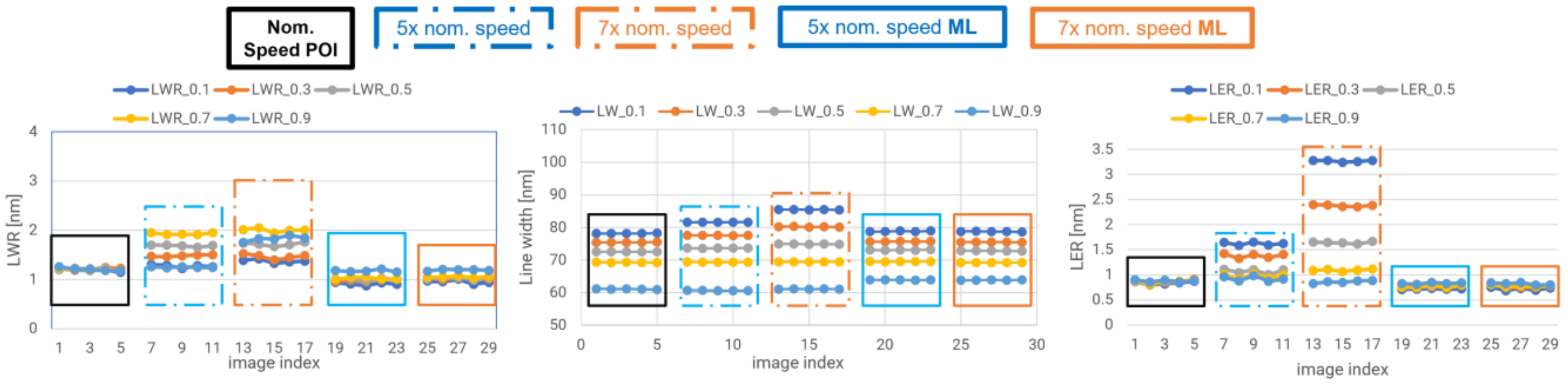

**Figure 8** Graphs showing the recovery of the accuracy and precision of the extracted lateral parameters after the application of transfer learning to the ML model needed for resolution enhancement on Line/Space structures of 100 nm pitch

## CONCLUSIONS

Metrology and inspection tools used in semiconductor fabs face a continuous push for increased throughput especially as we enter the age of massive metrology. While traditional SPM approaches have been enhanced with the introduction of high-through SPM tools into fabs, increasing scanning speeds beyond optimal levels results in a loss of accuracy and precision of captured images and the extracted parameters, especially in the lateral directions. In this work, we showcase a Res-UNet framework-based ML model for image resolution enhancement and provide results from real images captured at different speeds and used as inputs to the trained model. The implementation of this model enables the recovery of the fine details of a high-resolution image and critically allows for a return of the accuracy of the parameters monitored in process control. We additionally utilize transfer learning to reduce the data set and training period required for training a new model to be applied to structures of different dimensions and show that the results are as per expectations in this scenario as well. With the recent explosion in the utilization of ML in semiconductor fabs, this work provides inputs for the application of these models for further increasing the throughput of SPM-based metrology tools.

## ACKNOWLEDGEMENTS

Co-funded by The European Union. The project is supported by Key Digital Technologies and by RVO (Rijksdienst voor Ondernemend, RVO, The Netherlands). Views and opinions expressed are however those by the authors and do not necessarily reflect those of Key Digital Technologies joint undertaking. Neither the European Union nor the granting authority can be held responsible for them.